\pdfoutput=1

\documentclass[11pt]{article}

\usepackage[preprint]{acl}

\usepackage{times}
\usepackage{latexsym}

\usepackage[T1]{fontenc}

\usepackage[utf8]{inputenc}

\usepackage{microtype}

\usepackage{inconsolata}

\usepackage{graphicx}
\usepackage{algorithm}
\usepackage{algpseudocode}
\usepackage{microtype}
\usepackage{graphicx}
\usepackage{subfigure}
\usepackage{booktabs} 
\usepackage{pifont}
\usepackage{MnSymbol}
\usepackage{colortbl}

\definecolor{ourblue1}{RGB}{92,128,165}
\definecolor{ourblue2}{RGB}{99,142,180}
\definecolor{ourblue3}{RGB}{163,193,212}
\definecolor{ourblue4}{RGB}{214,224,238}

\title{Time Awareness in Large Language Models: \\Benchmarking Fact Recall Across Time}

\author{David Herel \\
  FEE, CTU in Prague \\
  CIIRC, Czech Technical University \\
  \texttt{hereldav@fel.cvut.cz} \\\And
  Vojtech Bartek \\
  FEE, CTU in Prague \\
  \texttt{bartevoj@fel.cvut.cz} \\\AND
    Jiri Jirak \\
  FEE, CTU in Prague \\
  \texttt{jirakji1@fel.cvut.cz} \\\And
  Tomas Mikolov \\
  CIIRC, Czech Technical University in Prague \\
  \texttt{tmikolov@gmail.com} \\
  }

\begin{document}
\maketitle
\begin{abstract}
Who is the US President? The answer changes depending on when the question is asked. While large language models (LLMs) are evaluated on various reasoning tasks, they often miss a crucial dimension: time. In real-world scenarios, the correctness of answers is frequently tied to temporal context. To address this gap, we present a novel framework and dataset spanning over 8,000 events from 2018 to 2024, annotated with day-level granularity and sourced globally across domains such as politics, science, and business. 
Our \emph{TimeShift} evaluation method systematically probes LLMs for temporal reasoning, revealing that base models often outperform instruction-tuned and synthetic-trained counterparts on time-sensitive recall. Additionally, we find that even large-scale models exhibit brittleness in handling paraphrased facts, highlighting unresolved challenges in temporal consistency. By identifying these limitations, our work provides a significant step toward advancing time-aware language models capable of adapting to the dynamic nature of real-world knowledge.
\end{abstract}

\section{Introduction}
\begin{figure}[h]
\centering
\includegraphics[width=\linewidth]{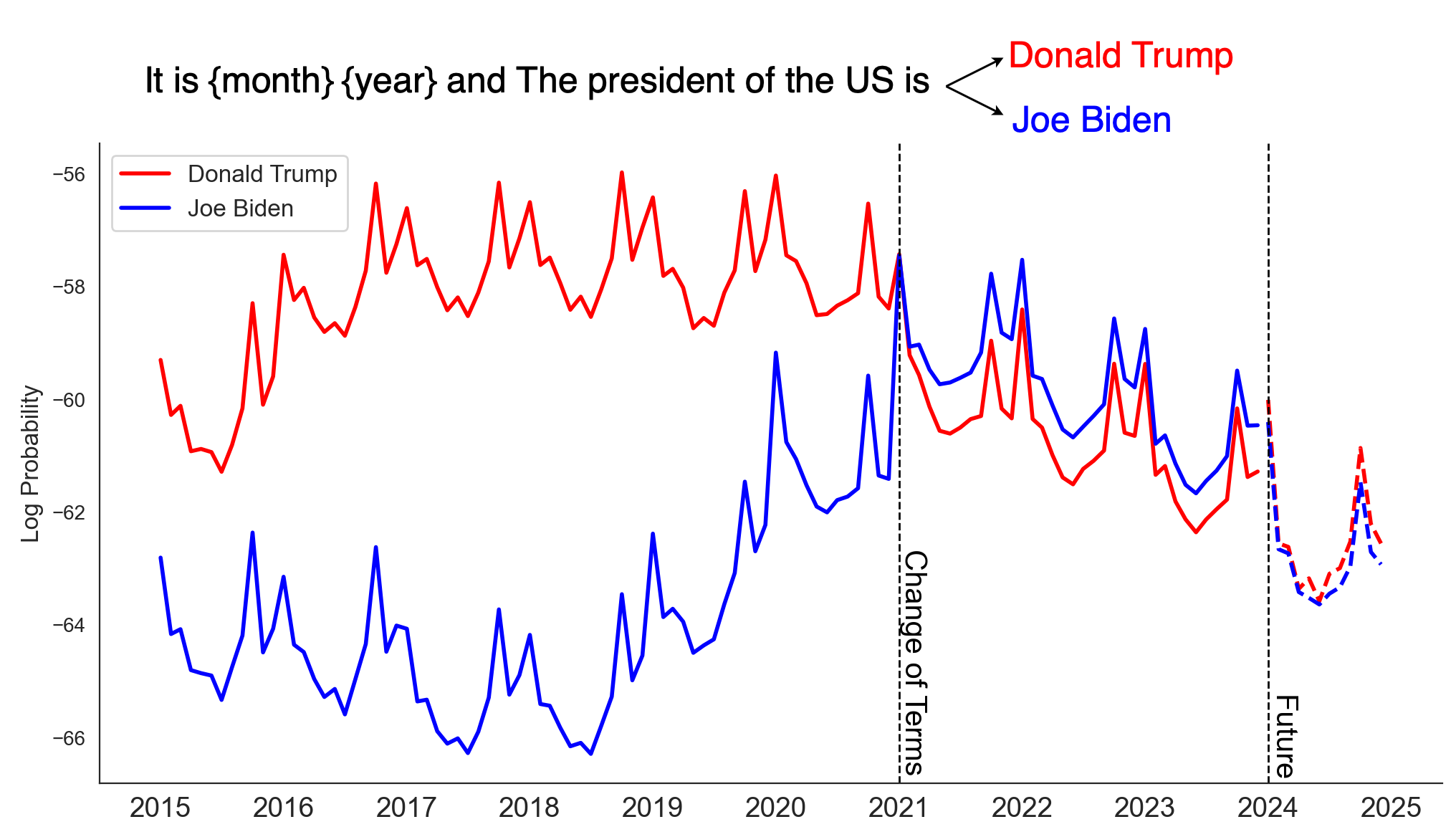}
\caption{Temporal log probabilities of sentences predicting the U.S. president (Joe Biden or Donald Trump) using Llama 3.2 3B, showing a clear shift in predictions aligned with their terms. As the model's training data cuts off at the end of 2023, predictions beyond this point reflect extrapolated trends.}
\label{fig:first_page_example}
\end{figure}

Large language models (LLMs) have revolutionized natural language understanding, reasoning, and factual recall, becoming foundational tools for applications such as chat bots \cite{brown2020languagemodelsfewshotlearners, openai2024gpt4technicalreport, touvron2023llamaopenefficientfoundation}, search engines \cite{thakur2021beirheterogenousbenchmarkzeroshot}, and automated fact-checkers \cite{petroni2019languagemodelsknowledgebases, roberts2020knowledgepackparameterslanguage}. However, their ability to handle time-sensitive facts—a critical component of real-world knowledge—remains under-explored. In many scenarios, the correctness of an answer depends not only on the question but also on when it is asked. For example, “Who is the US President on November 9, 2020, versus January 21, 2021?” requires reasoning tied to specific dates, a capability that current benchmarks often overlook.

Time awareness is crucial for dynamic tasks such as real-time fact-checking, knowledge base maintenance, and temporal question answering. While LLMs excel at static factual recall and general reasoning, their performance on time-dependent queries remains an open challenge. To address this, our approach systematically probes models for temporal reasoning by measuring the log probabilities of time-sensitive sentences across different temporal contexts. For example, we evaluate whether the log probabilities of sentences like “Donald Trump is the US president” and “Joe Biden is the US president” shift appropriately as leadership changes over time. As illustrated in Figure \ref{fig:first_page_example}, our approach captures these temporal dynamics, with models like Llama 3.1 8B \cite{llama3} showing partial success in adjusting predictions based on temporal prefixes. This highlights the importance of fine-grained temporal evaluation, which current benchmarks fail to capture comprehensively.

To address this gap, we introduce a novel dataset and evaluation framework designed to rigorously test daily temporal awareness in LLMs. Our dataset spans over 8,000 events from 2018 to 2024, annotated with day-level granularity and sourced globally across diverse domains such as politics, science, and business. Each event is paired with paraphrases to evaluate robustness in fact recall when phrasing varies. Using our \emph{TimeShift} evaluation method, we systematically probe models by generating temporal variations to assess their ability to reason across time and paraphrased contexts. 

Table \ref{tab:dataset_examples} highlights examples from our dataset, showcasing the diversity of events and annotations. This fine-grained, systematic approach allows us to uncover limitations in temporal reasoning across model families, including instruction-tuned models and synthetic-trained architectures. 

Our contributions are summarized as follows:
\begin{itemize}
    \item We introduce a comprehensive dataset with over 8,000 events spanning seven years, annotated with day-level granularity and paired with paraphrases, enabling robust evaluation of time-sensitive fact recall.
    \item We propose \emph{TimeShift}, a novel evaluation framework that systematically probes models’ temporal reasoning capabilities, uncovering key limitations in handling time-dependent queries.
    \item We provide a detailed evaluation of over a dozen state-of-the-art open-source LLMs, revealing that base models often outperform instruction-tuned and synthetic-trained models. Surprisingly, even large models exhibit brittleness when paraphrased facts are tested.
    \item All data, code, and evaluation tools are open-sourced to encourage further research into temporal reasoning in LLMs.
\end{itemize}

By addressing a critical gap in current benchmarks, this work lays the groundwork for advancing time-aware LLMs capable of reasoning about the dynamic nature of real-world knowledge.

\section{Related Work}
Several datasets have been introduced to evaluate the temporal reasoning capabilities of LLMs. The \textit{TempReason} dataset \cite{tan-etal-2023-towards} and \textit{TRAM} benchmark \cite{wang2024trambenchmarkingtemporalreasoning} both focus on assessing LLMs’ understanding of event order, duration, and frequency. However, these benchmarks primarily target broader temporal reasoning tasks rather than specific factual recall at finer time resolutions, such as determining the exact month when an event occurred.

An alternative approach involves modifying the self-attention mechanism \cite{vaswani2023attentionneed} to incorporate temporal information \cite{rosin-radinsky-2022-temporal}, improving performance on semantic change detection tasks \cite{schlechtweg-etal-2020-semeval, hamilton2018diachronicwordembeddingsreveal}. However, these adaptations have not been evaluated for their ability to recall specific temporal facts.

In addition, the \textit{TempLAMA} dataset \cite{Dhingra_2022} probes LLMs on facts associated with specific years but does not extend to the month or day-level precision required for many real-world applications. Similarly, the \textit{Test of Time} benchmark \cite{fatemi2024testtimebenchmarkevaluating} explores event relationships over time but lacks the focus on precise, time-bound factual recall.

\begin{table*}[ht!]
\small
\centering
\begin{tabular}{p{3.3cm}|p{3.3cm}|c|c|c|c|c|c|c}
\toprule
Original Sentence & Paraphrase 1 & ... & ... & ... & Year & Month & Day & Category \\ \midrule
Rolling Stone magazine co-founder Jann Wenner... & Jann Wenner, co-founder of... & ... & ... & ... & 2023 & 9 & 16 & Entertainment \& Arts \\ \hline
Meta launches Threads - Instagram's new... & Meta introduces Threads, a new app.." & ... & ... & ... & 2023 & 7 & 5 & Science \& Technology \\ \bottomrule
\end{tabular}
\caption{Examples from our dataset containing over 8,000 events with precise timestamps and paraphrases. For clarity, we display only a subset of paraphrases, omitting some metadata (country, source URLs) from this table.}
\label{tab:dataset_examples}
\end{table*}

\section{Dataset}
\begin{figure}[h!]
\centering
\includegraphics[width=0.46\textwidth]{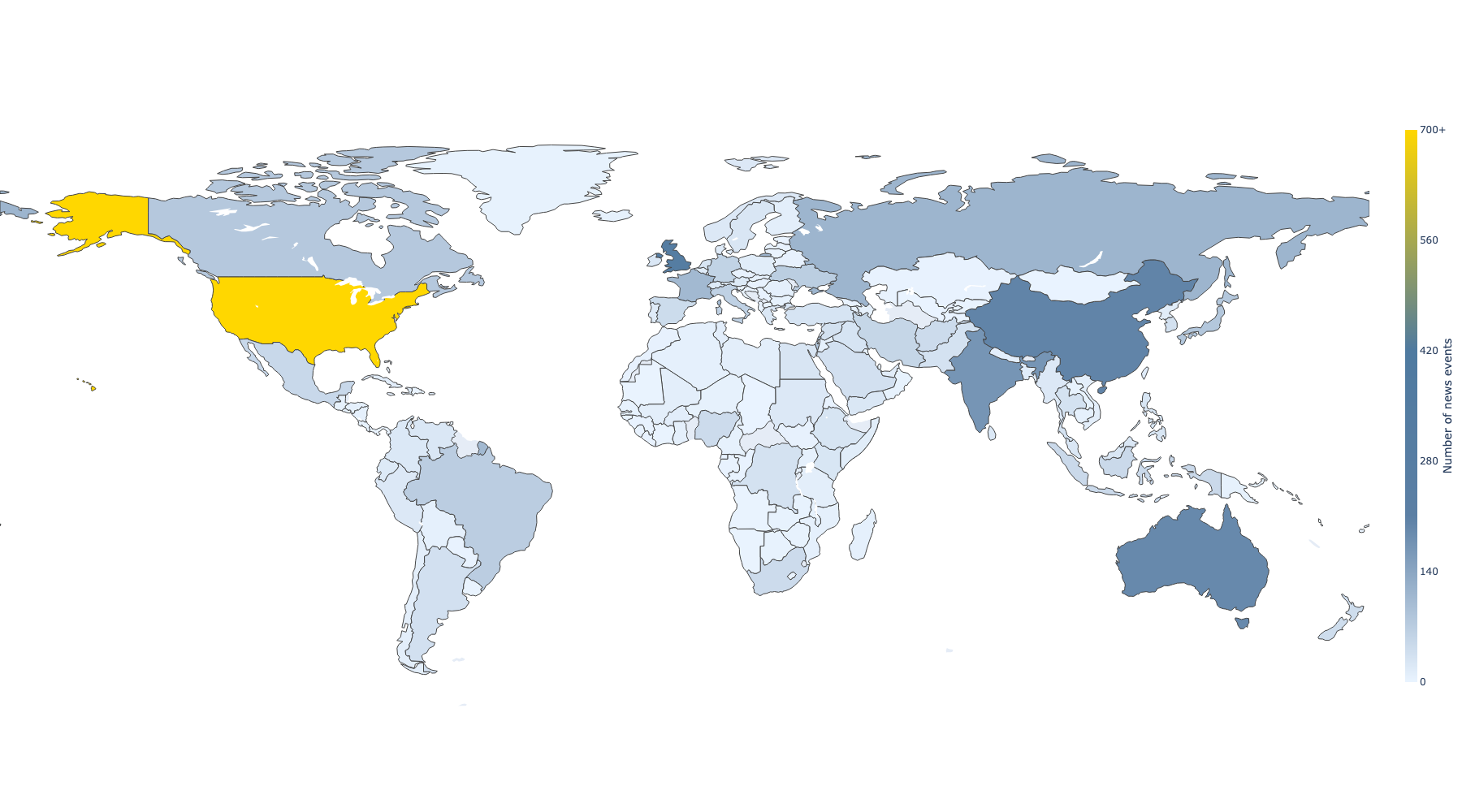}
\caption{World map showing the amount of news per country, US is in the first place with over 3,700 events across the 7 years.}
\label{fig:world_map}
\end{figure}

Our dataset is designed to assess LLMs' temporal awareness, specifically their ability to recall facts tied to specific dates. It comprises over 8,000 significant events from 2018 to 2024 across politics, business, science, art, and crime, ensuring geographical and cultural diversity. As an English-language dataset, geographically the highest event concentration is in the United Stated (3,700\texttt{+}), followed by global ($\approx950$) and UK ($\approx330$) as illustrated in Figure \ref{fig:world_map}.


Events are evenly distributed across months and days, though seasonal variations exist (e.g., increased reporting in summer, slight weekend decline). Each event is concisely represented by a headline of no more than 30 words, ensuring clarity and brevity, and is sourced from reputable and authoritative outlets (Section \ref{sec:data_collection}) to ensure accuracy and credibility.

\vspace{-5pt}

\subsection{Data Collection and Structure} \label{sec:data_collection}
The dataset was constructed by employing a custom web-scraping pipeline that extracted headlines from major global news outlets (e.g., BBC \cite{bbc_news2023}, Reuters \cite{reuters2023}, The New York Times \cite{nyt2023}), academic journals (e.g., Nature \cite{nature2022}), and government publications (e.g., official government websites, United Nations reports \cite{un_report2022}). To ensure accuracy, automated filtering mechanisms cross-referenced timestamps and removed duplicates, while heuristic-based checks discarded ambiguous events lacking clear temporal markers. Events with conflicting date information across sources were excluded to maintain consistency.

Each event in the dataset is annotated with its exact day, month, and year and is accompanied by four paraphrased versions. These paraphrases were generated through a combination of text transformation models and cross-source comparisons, ensuring variation in expression while preserving factual accuracy and similar length distribution (Appendix \ref{sec:lengthdist}). This variation is essential for evaluating the robustness of LLMs in factual recall when events are expressed differently. The dataset is specifically designed to assess whether models can recognize events despite rewording. Table \ref{tab:dataset_examples} provides an example of the dataset structure.

To categorize events, metadata tags were extracted during the scraping. If not available, we used a lightweight LLM-based classifier trained on labeled event data to infer these attributes.

By employing rigorous filtering, multi-source validation, and LLM-assisted classification, our dataset provides a high-fidelity benchmark for evaluating LLMs' ability to recall time-sensitive facts with precision.

\subsection{Category and Temporal Distribution}

\begin{figure}[h!]
\centering
\includegraphics[width=0.45\textwidth]{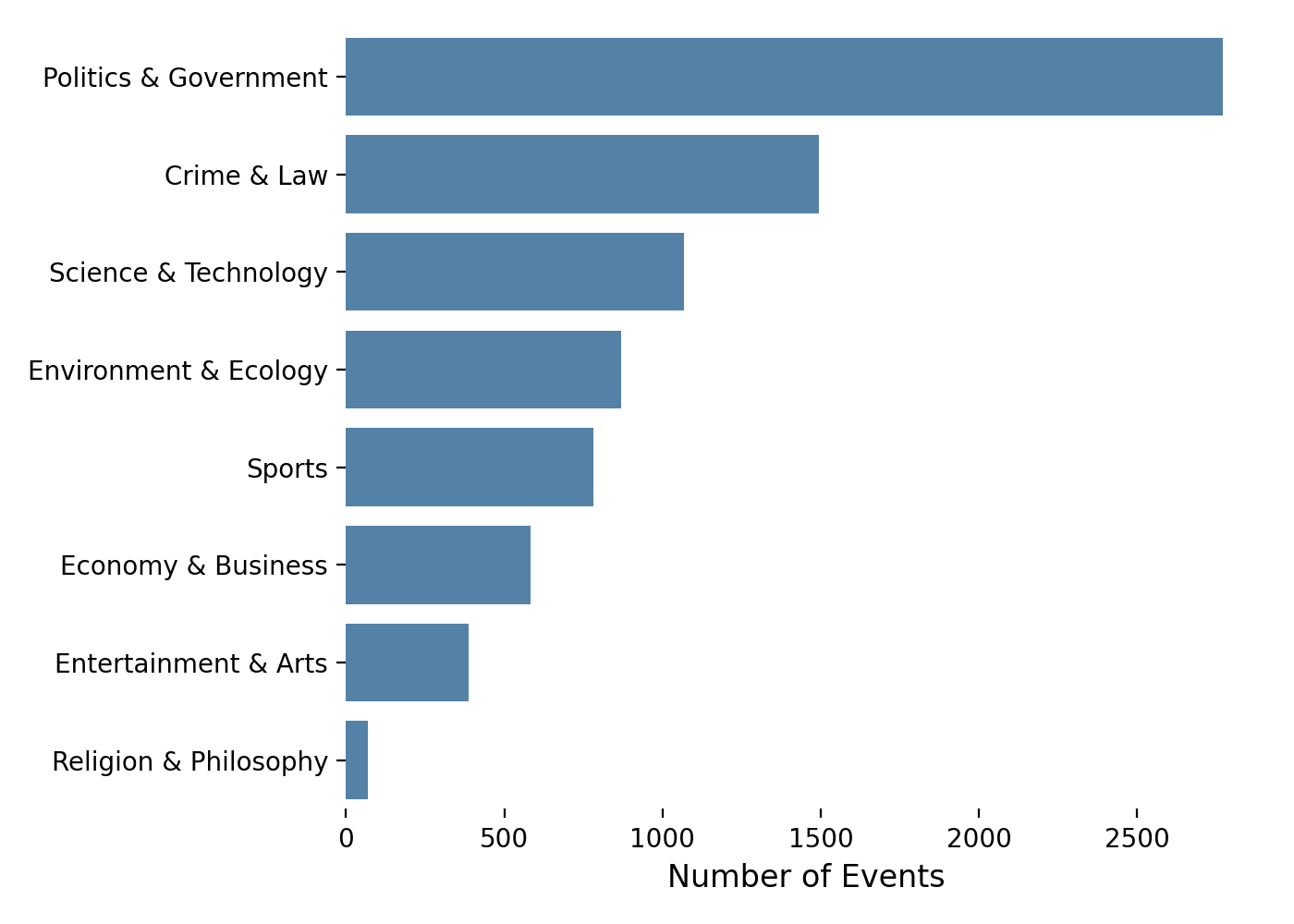}
\vspace{-5pt}
\caption{Distribution of events across categories, showing the highest concentration in Politics \& Government and Crime \& Law categories.}
\label{fig:category_distribution_graphs}
\end{figure}

\begin{figure}[h!]
\centering
\includegraphics[width=0.45\textwidth]{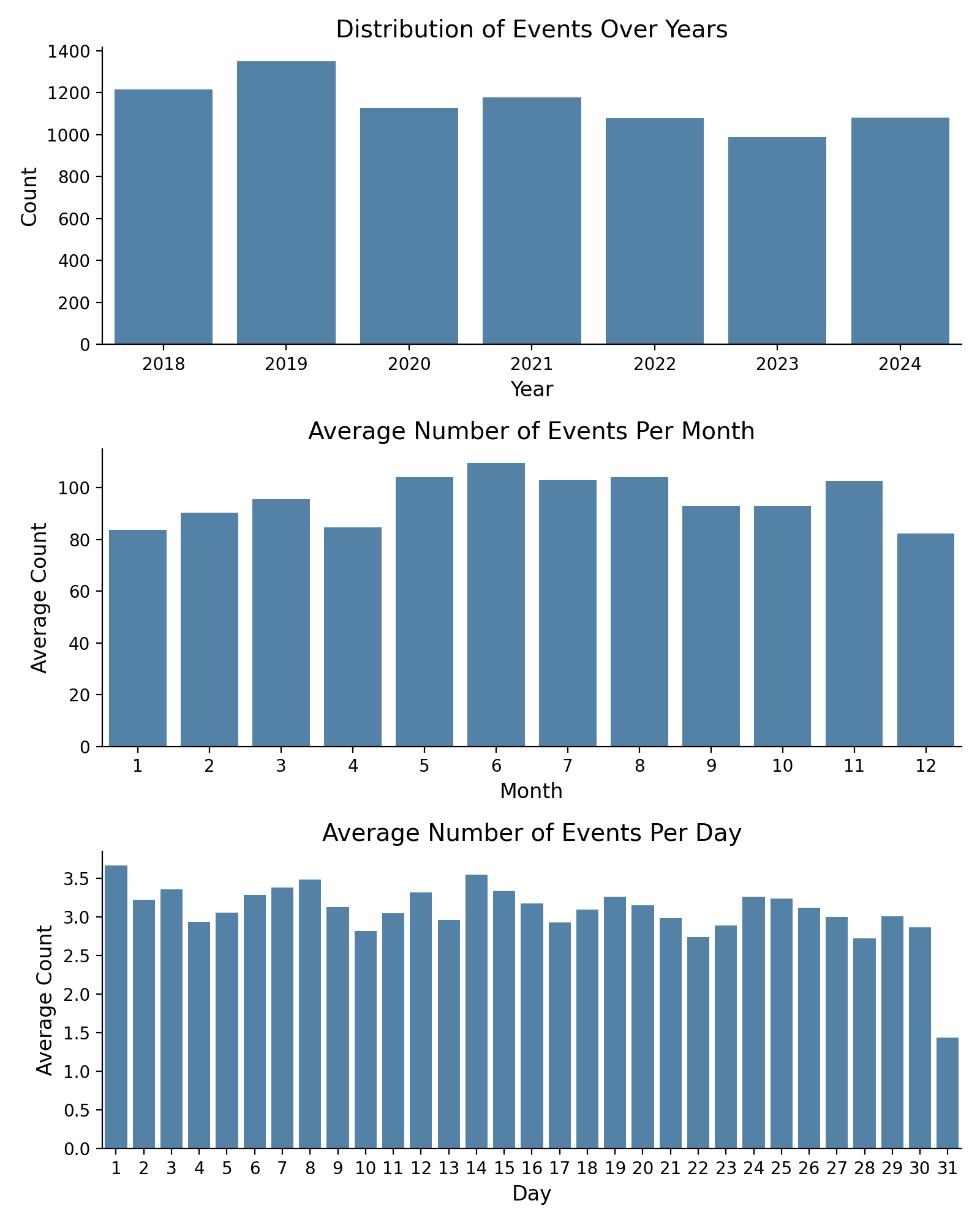}
\vspace{-5pt}
\caption{Even distribution of events across years, months, and days, ensuring balanced temporal coverage for evaluation.}
\label{fig:time_distribution_graphs}
\end{figure}
\vspace{-5pt}

The dataset spans a diverse range of categories, as illustrated in Figure \ref{fig:category_distribution_graphs}. On average, each day includes approximately three events, with some seasonal variations—such as a slight increase during summer months and a decline on weekends. The temporal distribution across years, months, and days is shown in Figure \ref{fig:time_distribution_graphs}, ensuring a balanced representation that prevents any specific period from disproportionately affecting the results.


\begin{table*}[ht!]
\centering
\small
\begin{tabular}{l|c|c|c|c}
\toprule
\textbf{Prefix Format} & \textbf{Year Acc.} & \textbf{Month Acc.} & \textbf{Day Acc.} & \textbf{Final Acc.} \\ 
\midrule
\texttt{It is \{day\} \{month\}, \{year\}. \{event\}} & 31.5\% & 4.9\% & 0.4\% & \textbf{12.3\%} \\  
\texttt{It is \{month\} \{day\}, \{year\}. \{event\}} & 31.5\% & 4.9\% & 0.2\% & 12.2\% \\  
\texttt{It is \{year\} \{month\}, \{day\}. \{event\}} & 31.5\% & 2.5\% & 0.2\% & 11.4\% \\  
\texttt{It is \{year\} \{month\}, \{day\} and \{event\}} & 31.5\% & 2.5\% & 0.0\% & 11.3\% \\  
\texttt{\{day\}.\{month\}.\{year\}, \{event\}} & 29.6\% & 3.4\% & 0.2\% & 11.1\% \\  
\texttt{On \{month\} \{day\}, \{year\}, \{event\}} & 26.9\% & 4.4\% & 0.5\% & 10.6\% \\  
\texttt{On \{day\}/\{month\}/\{year\}, \{event\}} & 26.9\% & 4.4\% & 0.1\% & 10.5\% \\  
\texttt{\{year\}-\{month\}-\{day\}: \{event\}} & 26.7\% & 3.2\% & 0.0\% & 10.0\% \\  
\bottomrule
\end{tabular}
\caption{Comparison of selected date-prefix formats based on accuracy in predicting time-sensitive facts. We tested a wide range of prefixes and report the best-performing ones.}
\label{tab:prefix_selection}
\end{table*}

\subsection{Public Availability}

The dataset, along with the evaluation framework, is publicly available on HuggingFace and GitHub, providing the research community with an accessible resource to further explore time-sensitive fact recall in LLMs.\footnote{\url{https://huggingface.co/datasets/hereldav/TimeAware/}}

\section{Experiments} \label{Experiments}
The core hypothesis driving our dataset is that an LLM should assign the highest probability to the sentence describing an event with the correct temporal prefix—specifically, the day, month, and year in which the event occurred. This hypothesis underpins the evaluation setup, where the model is tested on its ability to select the correct temporal context from a range of possibilities.

For example, consider the sentence: "It is April 13, 2022. Rolling Stone magazine co-founder Jann Wenner..." Here, the temporal prefix ("It is April 13, 2022.") explicitly situates the event within a specific timeframe, providing a clear basis for the model’s probabilistic assessment.
This specific prefix was selected based on additional experiments in Section \ref{sec:prefix_selection}, where it best aligned predictions with temporal context.

\subsection{Prefix selection} \label{sec:prefix_selection}
Selecting the optimal prefix for evaluating temporal awareness in LLMs is crucial, as phrasing affects how models interpret time-sensitive queries. To identify the best-performing prefix, we tested various prefix formulations on 10\% of the dataset using Llama-3.2 1B, Llama-3.2 3B, and Gemma-2 2B. These models, spanning different parameter sizes and architectures, provided a representative assessment of prefix impact on the performance. We explored variations in word order (e.g., year-first, day-first), separators (e.g., commas, dashes, slashes), and explicit prepositions (e.g., “On {date}” vs. “It is {date}”).

From this extensive search, Table \ref{tab:prefix_selection} reports the best-performing prefixes. The highest final accuracy of 12.3\% was achieved with “It is \{day\} \{month\}, \{year\}. \{event\}”, making it the optimal choice for probing LLMs’ temporal recall. Notably, prefixes starting with the year (e.g., “It is {year} {month}, {day} and {event}”) reduced accuracy, suggesting models overemphasized the year while struggling with finer details. Similarly, while numerical date formats using separators (e.g., ”{day}.{month}.{year}, {event}” or ”{year}-{month}-{day}: {event}”) performed reasonably, they exhibited slightly lower day-level accuracy. Based on these findings, we adopt the top-performing prefix across all subsequent experiments to ensure reliable and consistent temporal evaluation of LLMs.

\subsection{TimeShift Algorithm}

\begin{figure}[h!]
\centering
\includegraphics[width=0.5\textwidth]{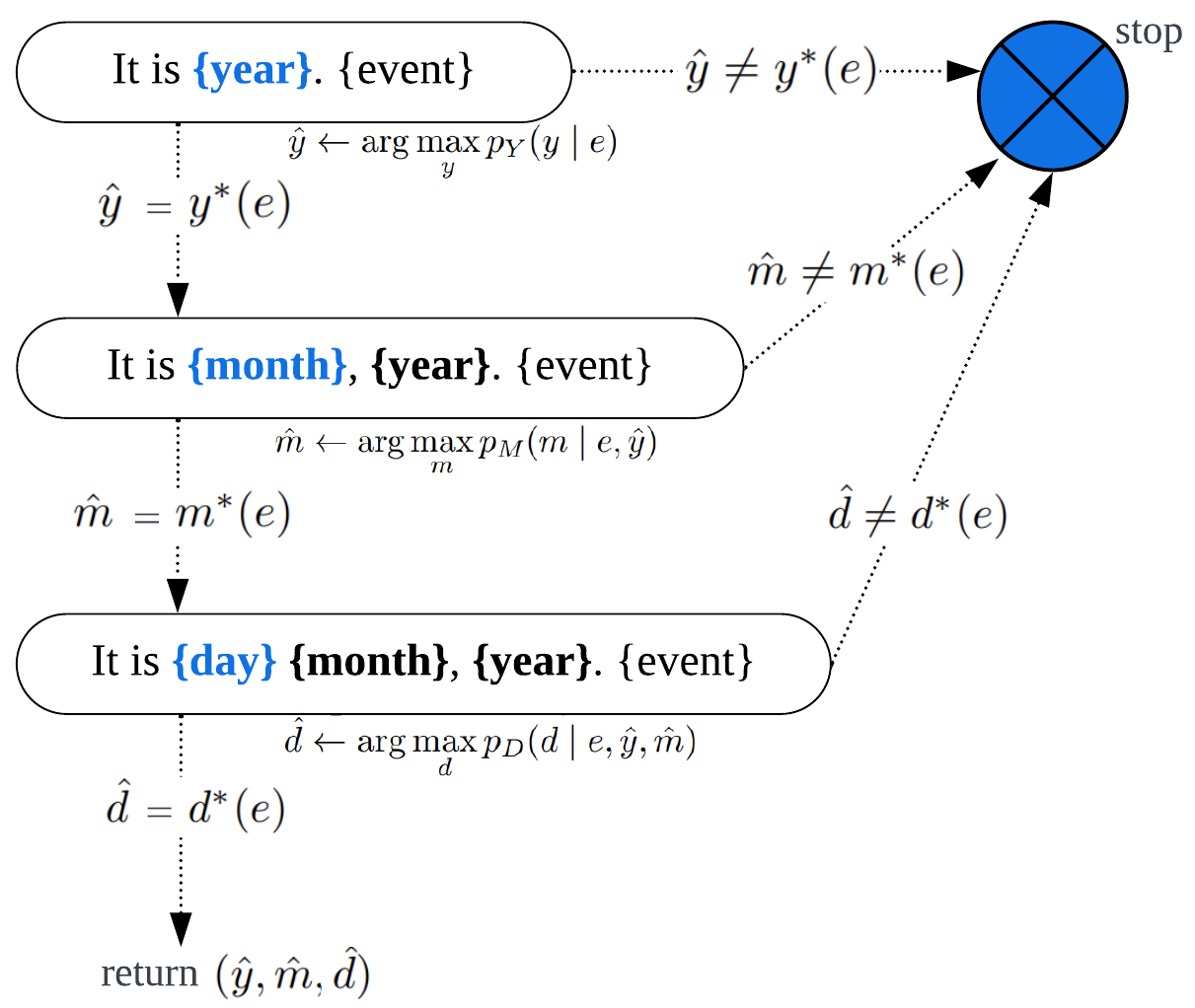}
\caption{Schema of the \emph{TimeShift} algorithm. Nodes represent sentences for which probabilities are computed with varying temporal prefixes (in blue). The sentence with the highest probability is selected as the prediction.}
\label{fig:algoschema}
\end{figure}

The \emph{TimeShift} algorithm evaluates a model’s ability to correctly predict the time of occurrence of an event. Given an event \( e \), our goal is to predict its correct date, structured as:

\begin{itemize}
    \item \( y^*(e) \) — the true year of the event,
    \item \( m^*(e) \) — the true month of the event,
    \item \( d^*(e) \) — the true day of the event.
\end{itemize}

The model generates probability distributions, specifically representing the likelihood of the entire sentence, including the temporal prefix. To ensure numerical stability, we compute the sum of the logarithms of these probabilities for each time unit:

\begin{itemize}
    \item \( p_Y(y \mid e) \) — probability of the event occurring in year \( y \),
    \item \( p_M(m \mid e, \hat{y}) \) — probability of the event occurring in month \( m \), given the predicted year \( \hat{y} \),
    \item \( p_D(d \mid e, \hat{y}, \hat{m}) \) — probability of the event occurring on day \( d \), given the predicted year \( \hat{y} \) and month \( \hat{m} \).
\end{itemize}

Instead of evaluating all possible (year, month, day) combinations, we apply a sequential filtering approach as described in Algorithm \ref{alg:timeshift} and Schema \ref{fig:algoschema} improving efficiency while preserving accuracy.
\vspace{3pt}

\begin{algorithm}
\caption{TimeShift}
\label{alg:timeshift}
\begin{algorithmic}[1]
    \State \textbf{Input:} Event \( e \)
    \State \textbf{Output:} Predicted date \( (\hat{y}, \hat{m}, \hat{d}) \)

    \State \textbf{Step 1: Predict Year}
    \State \( \hat{y} \gets \arg\max\limits_{y} p_Y(y \mid e) \)
    \If{\( \hat{y} \neq y^*(e) \)}
        \State \textbf{Stop (Incorrect Year)}
    \EndIf

    \State \textbf{Step 2: Predict Month (If Year is Correct)}
    \State \( \hat{m} \gets \arg\max\limits_{m} p_M(m \mid e, \hat{y}) \)
    \If{\( \hat{m} \neq m^*(e) \)}
        \State \textbf{Stop (Incorrect Month)}
    \EndIf

    \State \textbf{Step 3: Predict Day (If Month is Correct)}
    \State \( \hat{d} \gets \arg\max\limits_{d} p_D(d \mid e, \hat{y}, \hat{m}) \)
    \If{\( \hat{d} \neq d^*(e) \)}
        \State \textbf{Stop (Incorrect Day)}
    \EndIf

    \State \textbf{Return} \( (\hat{y}, \hat{m}, \hat{d}) \)
\end{algorithmic}
\end{algorithm}

\subsection{Stability Measurement Algorithm}

The Stability Measurement algorithm evaluates the robustness of a model’s predictions under minor input variations. Instead of measuring absolute accuracy, it quantifies whether the model maintains consistent predictions when an event description is paraphrased.

Given an event \( e \), the dataset provides:

\begin{itemize}
    \item \textbf{Original sentence:} \( e \) (news event).
    \item \textbf{Paraphrased sentences:} \( e'_1, e'_2, e'_3, e'_4 \) (four paraphrases of \( e \)).
    \item \textbf{True label:} \( y^*(e) \) (ground-truth year, month, or day).
\end{itemize}

Given these inputs, the model produces:

\begin{itemize}    
    \item \textbf{Prediction for the original sentence:} \( \hat{y}(e) \).
    \item \textbf{Predictions for the paraphrased sentences:} \( \hat{y}(e'_i) \) for \( i \in \{1,2,3,4\} \).
\end{itemize}

We define the following probability metric:

\textbf{Stability Probability:}  
    The probability that the model predicts the same correct result for a paraphrased event, given that it was correct for the original event:
    \[
    S = P(\hat{y}(e'_i) = y^*(e) \mid \hat{y}(e) = y^*(e))
    \]

Here, \( S \) quantifies the stability of predictions under paraphrasing.  
A high \( S \) indicates that the model is robust to rewording, whereas a low \( S \) suggests that the model is sensitive to variations in input phrasing.

\vspace{5pt}

\subsection{Why Use Log Probabilities Instead of Direct QA?}
A natural alternative would be to directly prompt the model with an open-ended question, such as \textit{"When did this event occur?"}, and compare the generated response to the ground truth. However, model responses in a free-form QA setting are inherently non-deterministic—even at temperature zero, variations in rounding, tokenization, and parallelization can lead to inconsistencies. This makes direct QA evaluation difficult to reproduce.

Log probabilities offer a structured, reproducible alternative by ranking all possible dates for an event in a probabilistic framework. While one could attempt to rank QA-based responses (e.g., via likelihoods or multiple-choice scoring), such approaches introduce additional challenges, such as inconsistent specificity (e.g., "early 2023" vs. "January 5, 2023") and reliance on post-processing heuristics. Our method avoids these pitfalls by ensuring a consistent evaluation framework across models.

\vspace{5pt}
\subsection{Metrics} \label{sec:metrics}

We evaluate models using two key metrics:

\begin{itemize}
    \item \textbf{Accuracy}: Accuracy is evaluated at multiple temporal granularities:
    \begin{itemize}
        \item \textbf{Yearly Accuracy}: The probability that the model correctly predicts the year of an event.
        \item \textbf{Monthly Accuracy}: Computed only for instances where the year is correctly predicted. Since it is conditioned on a correct yearly prediction, its sample size is smaller.
        \item \textbf{Daily Accuracy}: Assessed only when both the year and month predictions are correct.
        \item \textbf{Approximate Daily Accuracy}: Similar to daily accuracy but allowing for a $\pm1$-day margin of error.
        \item \textbf{Total Daily Accuracy}: The overall probability of correctly predicting an event’s exact date, computed as the product of yearly, monthly, and daily accuracies.
    \end{itemize}

\vspace{2pt}
    \item \textbf{Stability}: Stability measures the model’s robustness to input paraphrasing, evaluating its consistency in predicting the correct date across reworded event descriptions. Stability is computed at the following levels:
    \begin{itemize}
        \item \textbf{Yearly Stability}: The probability that the model predicts the correct year for a paraphrased event, given that it was correctly predicted for the original event.
        \item \textbf{Monthly Stability}: Evaluated only for instances where the model correctly predicts the year.
        \item \textbf{Daily Stability}: Assessed when both the year and month predictions are correct.
        \item \textbf{Approximate Daily Stability}: Similar to daily stability but allowing for a $\pm1$-day margin of error.
    \end{itemize}
\end{itemize}

\section{Results}
Our experiments reveal several clear trends in LLM performance on the time-sensitive fact recall task. We evaluated base and instruction-tuned variants, with parameter sizes ranging from 1B to 72B. Accuracy results are presented in Table \ref{tab:acc}, while stability results are detailed in Table \ref{tab:stability}.

\begin{table*}[ht!]
\small
\centering
\begin{tabular}{l|c|c|c|c|c}
\toprule
\textbf{Models} & \textbf{Yearly Acc} & \textbf{Monthly Acc} & \textbf{Daily Acc} & \textbf{Daily Acc Approx} & \textbf{Daily Total} \\ \midrule
\rowcolor{ourblue4} Qwen-2.5 1.5B & 22.07\% & 11.98\% & 3.30\% & 10.38\% & 0.09\% \\ 
Qwen-2.5 1.5B & 23.48\% & 11.74\% & 3.62\% & 13.57\% & 0.10\% \\ 
\rowcolor{ourblue4} Llama-3.2 1B & 25.89\% & 11.61\% & 4.15\% & 5.81\% & 0.12\% \\ 
\rowcolor{ourblue4} Phi-3.5-mini 3.8B & 27.57\% & 14.16\% & 3.83\% & 9.90\% & 0.15\% \\ 
\rowcolor{ourblue4} Phi-3-mini 3.8B & 26.00\% & 15.16\% & 4.11\% & 10.76\% & 0.16\% \\ 
\rowcolor{ourblue4} Gemma-2 2B & 29.65\% & 14.34\% & 4.99\% & 13.20\% & 0.21\% \\ 
\rowcolor{ourblue4} Qwen-2.5 7B & 29.94\% & 14.45\% & 5.48\% & 12.68\% & 0.24\% \\ 
Phi-4 14B & 34.26\% & 18.70\% & 3.89\% & 10.70\% & 0.25\% \\ 
Qwen-2.5 7B & 32.28\% & 14.87\% & 5.45\% & 13.51\% & 0.26\% \\ 
Llama-3.2 1B & 33.35\% & 13.57\% & 6.06\% & 11.57\% & 0.27\% \\ 
\rowcolor{ourblue4} Llama-3.2 3B & 32.81\% & 17.75\% & 4.71\% & 14.56\% & 0.27\% \\ 
Gemma-2 2B & 35.24\% & 17.13\% & 5.58\% & 11.36\% & 0.34\% \\ 
Llama-3.2 3B & 40.86\% & 20.75\% & 4.12\% & 11.18\% & 0.35\% \\ 
\rowcolor{ourblue4} Gemma-2 9B & 40.11\% & 23.25\% & 4.68\% & 13.37\% & 0.44\% \\ 
Gemma-2 9B & 45.45\% & 24.28\% & 6.55\% & 15.82\% & 0.72\% \\ 
\rowcolor{ourblue4} Llama-3.1 8B & 46.92\% & 27.66\% & 5.67\% & 12.30\% & 0.74\% \\ 
Llama-3.1 8B & 48.47\% & 27.42\% & 5.72\% & 12.10\% & 0.76\% \\ 
\rowcolor{ourblue4} Mistral-Nemo 12B & 50.62\% & 31.70\% & 7.38\% & 14.69\% & 1.18\% \\ 
\rowcolor{ourblue4} Gemma-2 27B & 44.11\% & 35.98\% & 7.78\% & 17.75\% & 1.23\% \\ 
\rowcolor{ourblue4} Qwen-2.5 72B & 46.35\% & 32.66\% & 10.63\% & 21.66\% & 1.61\% \\ 
Qwen-2.5 72B & 48.47\% & 38.15\% & \textbf{11.19\%} & \textbf{22.59\%} & 2.07\% \\ 
Gemma-2 27B & \textbf{55.16\%} & \textbf{42.09\%} & 8.97\% & 18.58\% & \textbf{2.08\%} \\ 
\bottomrule
\end{tabular}
\caption{Performance of large language models on the time-awareness benchmark. The table reports accuracy metrics at different granularities: yearly, monthly, daily, and an approximate daily measure that allows for a ±1-day error margin. The daily total accuracy reflects the model's likelihood of correctly assigning an event to its exact day, as further detailed in Section \ref{sec:metrics}. The results highlight the performance gap between instruction-tuned models (rows highlighted in blue) and their non-tuned counterparts, as well as the general accuracy improvement with increasing model size—except for Qwen-2.5 72B, which underperforms relative to smaller models. Gemma-2 27B achieves the highest overall accuracy.}
\label{tab:acc}
\end{table*}

\begin{table*}[ht!]
\small
\centering
\begin{tabular}{l|c|c|c|c}
\toprule
\textbf{Models} & \textbf{Yearly stability} & \textbf{Monthly stability} & \textbf{Daily stability} & \textbf{Daily stability Approx} \\ \midrule
\rowcolor{ourblue4} Qwen-2.5 1.5B &  11.14\% & 0.93\% & 0.02\% & 0.02\% \\ 
\rowcolor{ourblue4} Phi-3-mini 3.8B &  12.31\% & 1.11\% & 0.02\% & 0.02\% \\ 
Qwen-2.5 1.5B &  13.16\% & 0.84\% & 0.03\% & 0.03\% \\ 
\rowcolor{ourblue4} Phi-3.5-mini 3.8B  & 14.72\% & 1.38\% & 0.03\% & 0.04\% \\ 
\rowcolor{ourblue4} Llama-3.2 1B &  16.34\% & 0.98\% & 0.03\% & 0.03\% \\ 
\rowcolor{ourblue4} Qwen-2.5 7B  & 17.99\% & 2.02\% & 0.07\% & 0.07\% \\ 
\rowcolor{ourblue4} Gemma-2 2B  & 18.75\% & 1.91\% & 0.07\% & 0.09\% \\ 
Gemma-2 2B  & 20.31\% & 2.33\% & 0.07\% & 0.07\% \\ 
Llama-3.2 1B  & 20.42\% & 1.95\% & 0.06\% & 0.07\% \\ 
\rowcolor{ourblue4} Phi-4 14B  & 20.92\% & 2.81\% & 0.07\% & 0.09\% \\ 
Qwen-2.5 7B  & 22.34\% & 3.40\% & 0.07\% & 0.09\% \\ 
\rowcolor{ourblue4} Llama-3.2 3B  & 22.60\% & 2.26\% & 0.04\% & 0.05\% \\ 
Llama-3.2 3B & 25.77\% & 3.72\% & 0.07\% & 0.07\% \\ 
\rowcolor{ourblue4} Gemma-2 9B  & 32.31\% & 9.02\% & 2.50\% & 2.78\% \\ 
\rowcolor{ourblue4} Llama-3.1 8B  & 33.05\% & 9.12\% & 2.61\% & 2.52\% \\ 
\rowcolor{ourblue4} Qwen-2.5 72B & 33.52\% & 9.12\% & 2.00\% & 2.00\% \\ 
Qwen-2.5 72B & 35.05\% & 9.29\% & 2.33\% & 2.37\% \\ 
\rowcolor{ourblue4} Gemma-2 27B  & 36.39\% & 9.02\% & 2.12\% & 2.52\% \\ 
\rowcolor{ourblue4} Mistral-Nemo 12B  & 37.16\% & 9.31\% & 2.72\% & 2.75\% \\ 
Llama-3.1 8B  & 38.17\% & 10.33\% & 2.69\% & 2.72\% \\ 
Gemma-2 9B  & 38.19\% & 10.67\% & 2.47\% & 2.91\% \\ 
Gemma-2 27B & \textbf{39.39\%} & \textbf{11.02\%} & \textbf{3.12\%} & \textbf{3.12\%} \\ 
\bottomrule
\end{tabular}
\caption{Model stability analysis. Again, the instruct-tuned models are highlighted in blue. The table illustrates a general trend of increasing stability as the number of parameters grows. It also highlights the challenges of prompting the model for finer-grained time horizons—rephrasing a sentence while expecting the model to predict the exact same year, month, and day as in the original phrasing proves to be particularly difficult.}
\label{tab:stability}
\end{table*}

\subsection{Instruction-Tuned Models Underperform}

Across all model families, instruction-tuned variants underperform compared to base models on this task. For instance, Gemma-27B achieves 55.16\% Yearly accuracy but drops to 44.11\% after instruction tuning. Similarly, Llama-3.1 8B outperforms its instruction-tuned counterpart, achieving 38.17\% versus 33.05\%.

We hypothesize that the broad generalization achieved during instruction tuning dilutes time-specific factual recall, prioritizing task flexibility over detailed temporal knowledge.

\subsection{Impact of Model Size on Performance}

Model size exhibits a strong correlation with performance on our time-awareness benchmark, with larger models consistently outperforming smaller ones across all metrics. However, Qwen-2.5 72B deviates from this trend, underperforming relative to its parameter count. For instance, Gemma-2 27B Base achieves a Yearly accuracy of 55.16\%, surpassing Gemma-2 9B (45.45\%) and Gemma-2 2B (35.24\%). This pattern aligns with broader findings that larger models more effectively capture nuanced information, including temporal dependencies. Qwen-2.5’s weaker performance may stem from differences in training data quality, suboptimal fine-tuning, or a focus on multilingual generalization over precise factual retrieval.

\subsection{Underperformance of Synthetic-Training Models}

Despite excelling in reasoning and generation, synthetic-trained models like the Phi family struggle with temporal recall. Phi-3-mini 3.8B achieves only 26.00\% Yearly accuracy, while the larger Phi-4 14B reaches 34.26\%. Notably, the 1B parameter Llama-3.2 nearly matches Phi-4 with 33.35\%, suggesting that increased parameter size alone cannot offset the limitations of synthetic data. This highlights a key weakness: synthetic datasets, though effective for general knowledge, often lack the real-world temporal grounding needed for accurate recall of time-sensitive facts.

\newpage
\subsection{Performance over the years}
\begin{figure}[h!]
\centering
\includegraphics[width=0.49\textwidth]{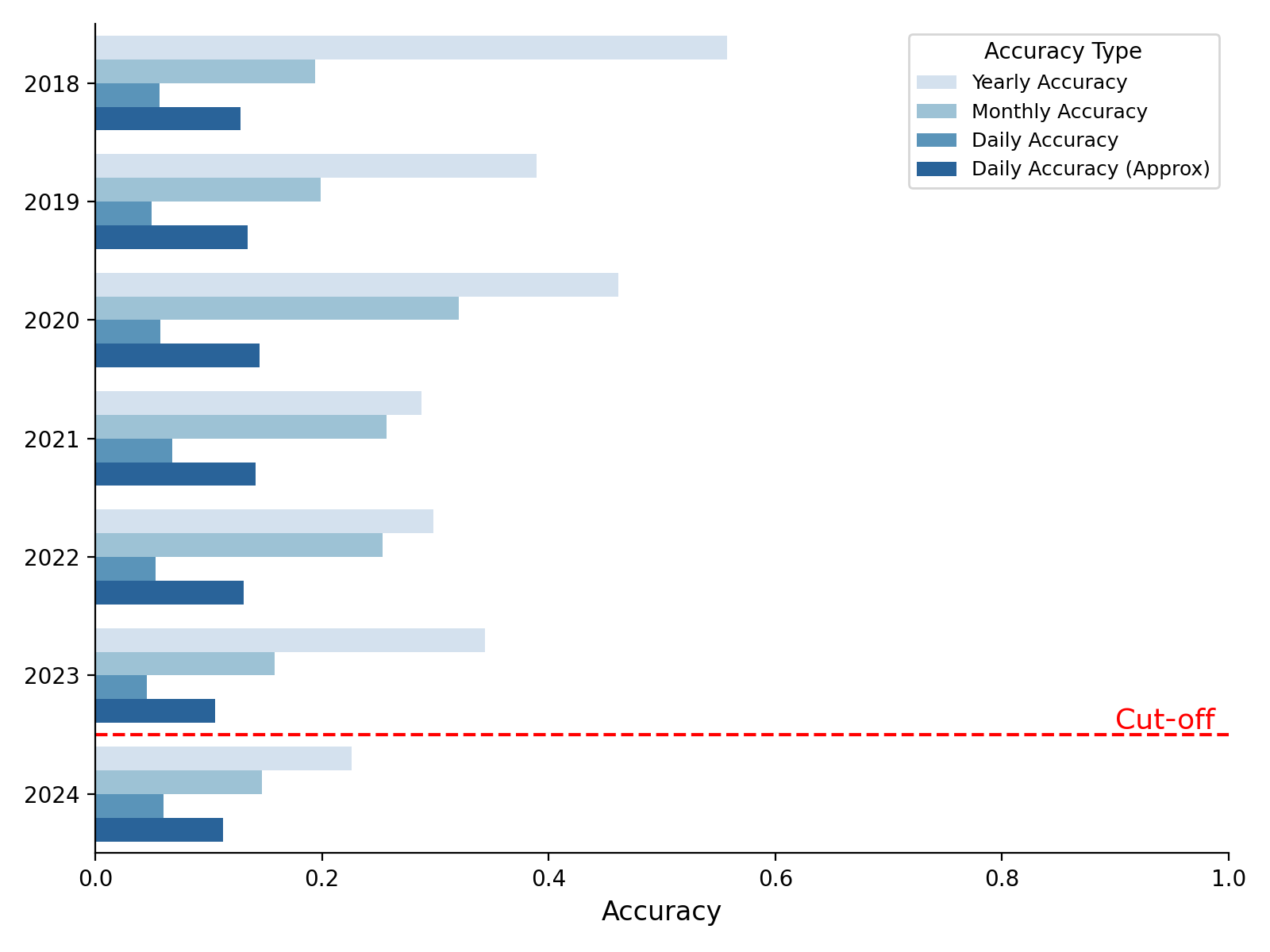}
\caption{Averaged performance of all models across years, showing a clear drop in performance as the cut-off date approaches.  }
\label{fig:year_acc}
\end{figure}

Nearly all models evaluated in this study were trained on datasets with a cut-off date of December 2023, as is explicitly stated for the Llama model family. For other models, which were released in the first half of 2024 but do not specify a precise cut-off date, it is reasonable to assume their training data extends to early 2024. This temporal boundary is clearly reflected in their performance, with a marked decline in accuracy for more recent facts, as illustrated in Figure \ref{fig:year_acc}. Interestingly, all models perform worse on recent events and better on older ones, likely due to the higher availability, stability, and reinforcement of past information in training data, whereas newer events are less represented and may undergo evolving interpretations.

\subsection{Stability Across Paraphrases}

Our stability evaluation highlights a significant susceptibility of models to variations in phrasing. Even the best-performing model in our study, Gemma-2 27B Base, correctly predicted the year in only 39.39\% of paraphrased cases, despite accurately classifying the original event. Smaller and instruction-tuned models, such as Llama-3.2 1B Base (20.42\%) and Qwen-2.5 1.5B Instruct (11.14\%), exhibited even greater sensitivity, frequently altering their predictions in response to minor rewording. This trend becomes even more pronounced as temporal granularity increases, with accuracy dropping to single-digit percentages for the best models in Daily stability assessments.

These findings reinforce a well-documented issue in LLMs: prompt sensitivity \cite{zhuo-etal-2024-prosa, sclar2024quantifyinglanguagemodelssensitivity, zhan2024unveilinglexicalsensitivityllms}. Larger models tend to demonstrate greater stability than their smaller counterparts, but even they struggle with consistency when faced with slight linguistic variations, emphasizing the ongoing challenge of robustness in factual recall.

\section{Conclusion}

In this paper, we introduced a novel dataset and evaluation benchmark specifically designed to assess LLMs' ability to handle time-sensitive facts, addressing a critical gap in existing evaluation methods that primarily focus on static factual recall. Our dataset, comprising over 8,000 events spanning from 2018 to 2024, provides a structured framework for testing models' temporal awareness by evaluating their ability to correctly associate events with their respective time periods.

Beyond assessing temporal accuracy, our dataset also enables the evaluation of model stability, offering insights into their robustness against variations in prompt phrasing. This aspect is crucial for understanding how reliably a model can maintain consistency in its predictions.

Our findings indicate that larger models consistently outperform smaller ones in time-sensitive tasks, reinforcing the role of scale in factual recall. However, instruction-tuned models, despite their strengths in general-purpose reasoning, struggle with temporal reasoning. Additionally, models trained primarily on synthetic data, such as those in the Phi family, demonstrate notable limitations in real-world temporal understanding. Furthermore, our analysis highlights the significant impact of prompt formulation on model behavior, revealing how slight variations in wording can lead to different predictions. 

Time awareness is essential for real-world applications such as virtual assistants, fact-checking, and temporal question-answering. By publicly releasing our dataset and evaluation framework, we aim to support further research in this area and encourage the community to extend this work.
\section*{Limitations}
While our benchmark provides a rigorous evaluation of time-sensitive factual recall, it has certain limitations. First, the dataset is exclusively in English, which may limit its applicability to multilingual LLMs. Future work should extend the dataset to include non-English events, enabling broader linguistic coverage. Second, our evaluation framework focuses on open-source models, as log probability access is required to perform a structured ranking of temporal claims. This constraint prevents us from directly assessing closed-source models, such as GPT-4 or Claude, unless they provide likelihood scores. Third, while our dataset covers diverse event categories and global sources, there remains an overrepresentation of Western events, particularly from the United States and the United Kingdom, due to the nature of English-language reporting. Expanding the dataset with multilingual and regionally balanced sources could mitigate this bias. 


\section*{Acknowledgments}
 The work was supported by the CTU internal grant SGS23/210/OHK3/3T/18 and by the Ministry of Education, Youth and Sports within the dedicated program ERC CZ under the project ARTEMIS no. LL2405."


\bibliography{main}

\appendix

\section{Appendix}
\label{sec:appendix}

In this section, we provide additional details on dataset properties, evaluation distributions, and experimental results. These supplementary figures further illustrate key aspects of our dataset's structure and the robustness of our evaluation framework.

\subsection{Paraphrase Length Distribution}\label{sec:lengthdist}

\begin{figure}[h]
\centering
\includegraphics[width=0.49\textwidth]{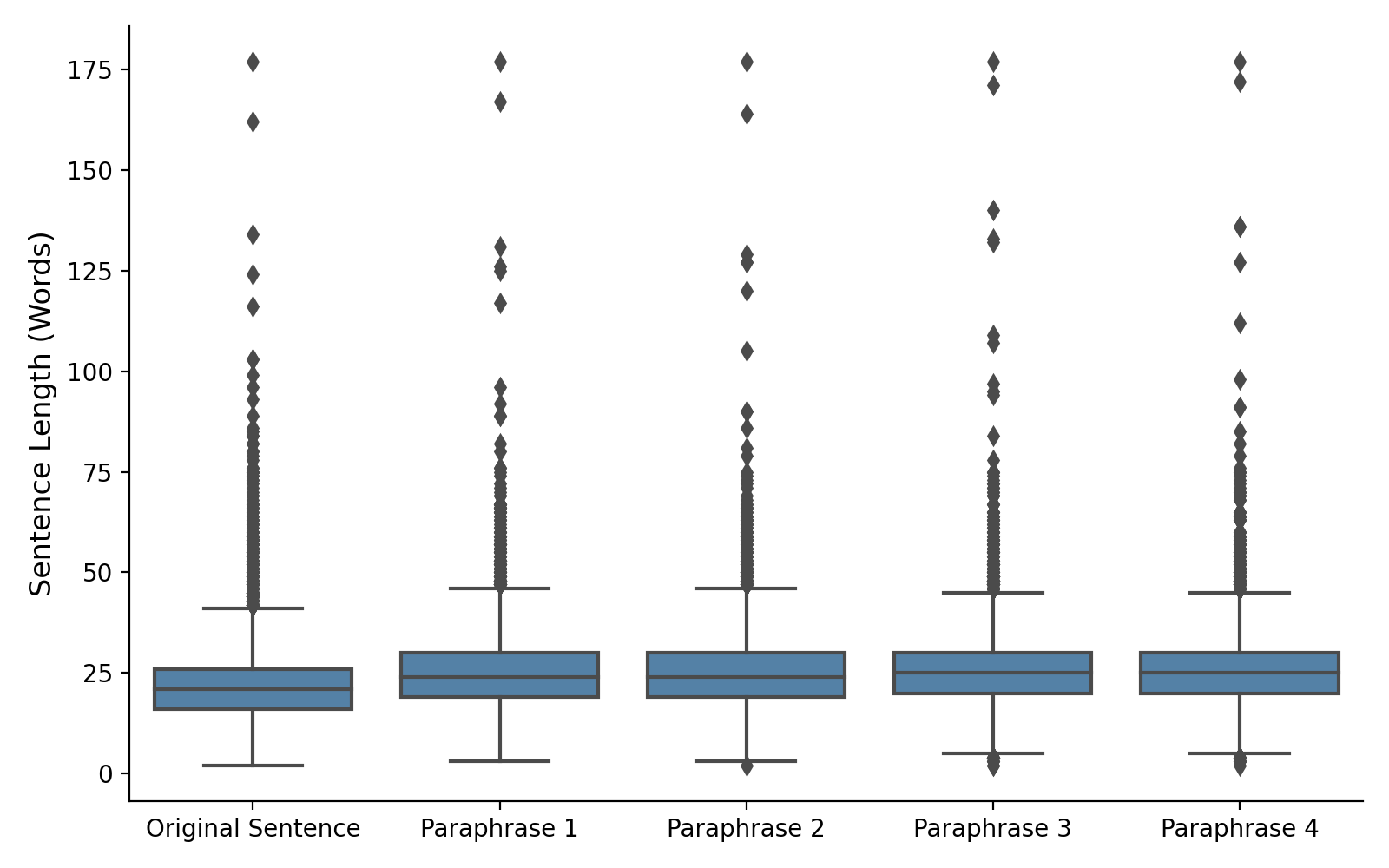}
\caption{Comparison of sentence length distributions between original event descriptions and their paraphrased counterparts. The alignment of distributions ensures that paraphrases do not introduce systematic biases in model evaluation.}
\label{fig:sentence_length_distribution}
\end{figure}

To ensure fair evaluation, we maintain similar length distributions between original event descriptions and their paraphrased variants. As shown in Figure~\ref{fig:sentence_length_distribution}, the paraphrased sentences closely follow the length distribution of the original events. This minimizes potential biases where longer or shorter phrasings could disproportionately affect model performance.

\subsection{Accuracy Across Event Categories}

\begin{figure}[h]
\centering
\includegraphics[width=0.49\textwidth]{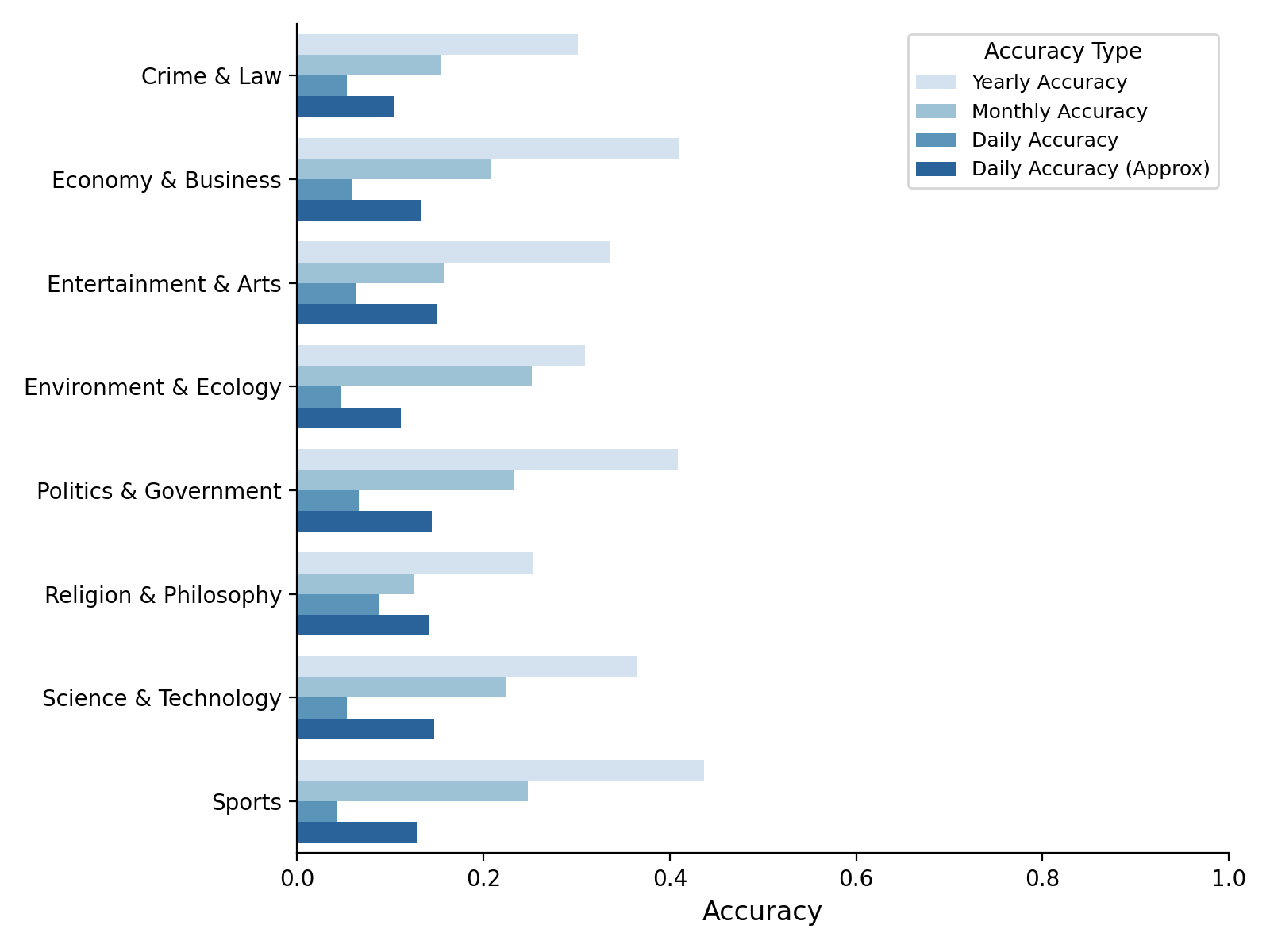}
\caption{Accuracy distribution across different event categories. The relatively uniform performance indicates that no single category disproportionately influences model accuracy, confirming the dataset’s balanced composition.}
\label{fig:categoryacc}
\end{figure}

To assess whether certain event categories are easier for models to predict, we analyze accuracy distributions across different domains. As depicted in Figure~\ref{fig:categoryacc}, the model performance remains relatively stable across categories such as politics, science, and entertainment. This suggests that the temporal reasoning task is not inherently skewed toward specific domains, reinforcing the general applicability of our benchmark.

\subsection{Accuracy Across Geographical Regions}

\begin{figure}[h]
\centering
\includegraphics[width=0.49\textwidth]{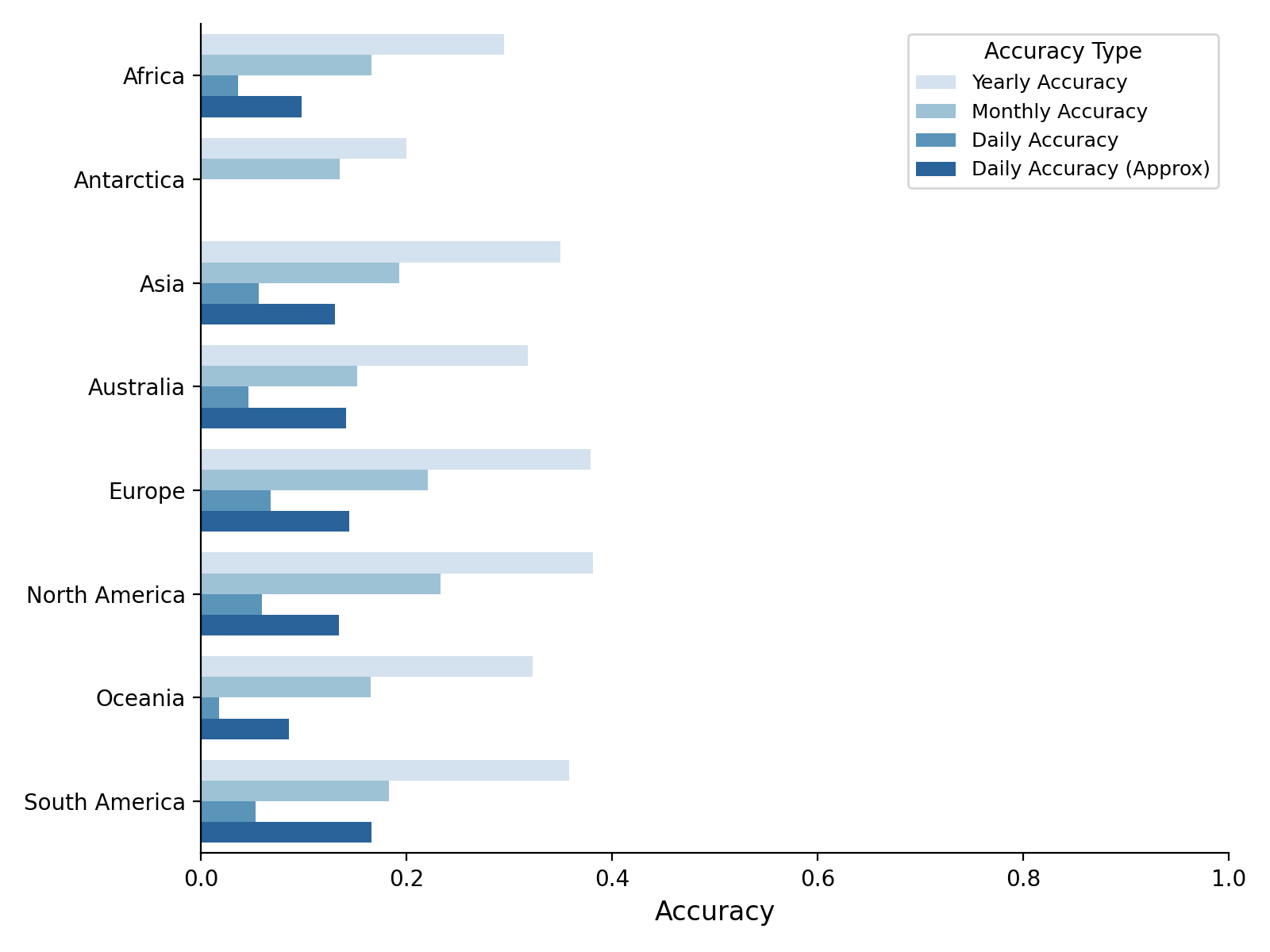}
\caption{Accuracy distribution across continents. The balanced accuracy levels suggest that the dataset provides a fair temporal reasoning challenge across diverse geographic regions, despite the English-language focus.}
\label{fig:continentacc}
\end{figure}

Similarly, we analyze model performance across different continents to ensure that temporal reasoning is not biased toward specific geographic regions. Figure~\ref{fig:continentacc} shows that accuracy is relatively consistent across continents, further supporting the robustness of our evaluation framework. While English-language news sources naturally introduce an overrepresentation of Western events (e.g., USA and UK), our dataset remains diverse enough to challenge models across a wide range of global events.

\subsection{Additional Insights and Future Work}

Overall, the even accuracy distribution across event categories and geographic regions reinforces the robustness of our dataset. The alignment of paraphrased sentence lengths with original event descriptions ensures that variations in sentence structure do not introduce unintended evaluation biases.

Future work could extend the dataset by incorporating multilingual event sources, enabling evaluation across different linguistic and cultural contexts. Additionally, refining the event selection pipeline to ensure better coverage of underrepresented regions could further enhance the dataset’s utility.

\end{document}